\documentclass[conference]{IEEEtran}

\usepackage{latexsym}
\usepackage{graphicx}
\usepackage{tikz}
\usepackage{booktabs}   
\usepackage{multirow}
\usepackage{url}
\usepackage{cite}
\usepackage{amsmath,amssymb,amsfonts}
\usepackage{algorithmic}
\usepackage{textcomp}
\usepackage{xcolor}
\usepackage{subcaption}
\usepackage{tabularx}
\usepackage{array}
\usepackage[hidelinks]{hyperref}
\usetikzlibrary{positioning,arrows.meta,fit,backgrounds}

\usepackage{siunitx}
\tikzset{
  textbox/.style={
    rectangle, rounded corners,
    minimum width=2cm, minimum height=2.2cm,
    text centered, align=center,
    draw=gray, fill=white, line width=0.7pt
  },
  llm/.style={
    rectangle, rounded corners,
    minimum width=2cm, minimum height=2.2cm,
    text centered, align=center,
    draw=black, fill=blue!20, line width=0.9pt,
  },
  topbox/.style={
    rectangle, rounded corners,
    minimum width=2cm, minimum height=1.5cm,
    text centered, align=center,
    draw=gray!70, fill=orange!12, line width=0.7pt
  },
  groupbox/.style={
    draw=gray!70, rounded corners,
    line width=0.7pt, fill=gray!6,
    inner sep=10pt 
  },
  arrow/.style={-Stealth, line width=0.7pt},
  dashedarrow/.style={-Stealth, dashed, line width=0.8pt}
}

\def\BibTeX{{\rm B\kern-.05em{\sc i\kern-.025em b}\kern-.08em
    T\kern-.1667em\lower.7ex\hbox{E}\kern-.125emX}}
\begin{document}

\title{V\={a}kQA: A Benchmark and Evaluation Study for Telugu Spoken Factoid Question Answering}
\author{
\IEEEauthorblockN{
Bhavana Akkiraju\IEEEauthorrefmark{1},
Ravi Sastry Kolluru\IEEEauthorrefmark{1},
Charan Devarakonda\IEEEauthorrefmark{1}
Srihari Bandarupalli\IEEEauthorrefmark{1},\\
Santosh Kesiraju\IEEEauthorrefmark{2},
Anil Kumar Vuppala\IEEEauthorrefmark{1}
}
\IEEEauthorblockA{
\IEEEauthorrefmark{1}International Institute of Information
Technology Hyderabad, India 
\IEEEauthorrefmark{2}Brno University of Technology,
Speech@FIT, Czechia
}
\IEEEauthorblockA{
\small
\{bhavana.akkiraju,kolluru.s,sricharan.d,
srihari.bandarupalli\}@research.iiit.ac.in\\
}
}
\maketitle

\begin{abstract}

Question answering has advanced rapidly with large language models, but 
predominantly for high-resource languages, in both text and spoken settings.
Spoken question answering (SQA) benchmark for Telugu remains unexplored, and the
reliability of automatic evaluation in this setting remains unquantified.
We introduce V\={a}kQA, a Telugu SQA benchmark of 2,001 factoid question-answer pairs
across six domains, with 2.53 hours of speech audio, bilingual transcriptions, 
and human-verified reference answers. We first validate evaluation methods against human 
judgments: Gemini-as-a-judge best approximates human ratings but is non-uniformly strict, 
while open-weight judges systematically penalize correct Telugu answers that differ in 
surface form from the reference. Using this validated setup, we benchmark proprietary
and open-weight models across input modality, language, and domain. We observe that Telugu
phrasing retains cultural specificity that is lost in translation, speech input
introduces phonetic confusions that alter question meaning, and cascaded
ASR-MT errors compound progressively. V\={a}kQA is publicly released.

\end{abstract}


\section{Introduction}
Spoken Question Answering (SQA) brings together speech understanding and question answering: the input is speech, and the system must produce a direct answer. Spoken input introduces acoustic and linguistic variability that text does not, affecting both recognition and downstream reasoning, and for low-resource languages such as Telugu this is compounded by the scarcity of annotated speech resources.
Question answering research has been shaped largely by English benchmarks such as SQuAD~\cite{squad}, Natural Questions \cite{naturalqa-inproc}, which inspired subsequent multilingual QA benchmarks such as XQuAD~\cite{artetxe2020cross}, MLQA~\cite{lewis2020mlqa} and MKQA~\cite{longpre2021mkqa} relying on translations from English sources. On the other hand, TyDi QA~\cite{tydi} consists of questions written by native speakers of 11 typologically diverse languages. Indic-language QA resources have grown in recent years~\cite{doddapaneni2023towards,indicqa2024,benqa2024}. While most of them are created with the help of native human annotators, some are based on translations either from English or Hindi to several other Indian languages~\cite{rohera2024indicquest}. For Telugu specifically, TeQuAD offers a substantial text QA dataset but no speech variability~\cite{vemula2022tequad}, leaving Telugu spoken QA largely unexplored.

Existing spoken QA work elsewhere reinforces the need for native audio: Spoken SQuAD~\cite{lee2018spoken} used TTS-synthesized speech over SQuAD passages, and ODSQA~\cite{odsqa} built a Chinese open-domain resource from read-speech, both remaining extractive; SD-QA~\cite{sdqa} extended this to a multi-dialect setting across five languages and 24 dialects but keeps the same passage-grounded task as TyDi QA; SpokenNativQA collected natural, human-recorded queries in Arabic and English, motivated by the fact that most SQA data is English-centric and synthetic~\cite{alam25_interspeech}; and ViSQA applied the same TTS synthesis as Spoken SQuAD to Vietnamese~\cite{visqa}. Together, these show that a spoken benchmark is not simply text with audio attached: how the audio is collected, whether a passage is required, and how transcription is handled all shape the errors models make. In this work, we present V\={a}kQA benchmark where the questions come directly from spoken Telugu interaction (quiz-style) rather than translation or synthesis. In addition, we also provide original transcriptions and English translations. Our work adds a systematic evaluation across input language, modality, cascaded ASR $\rightarrow$ MT errors, and judge reliability --- not attempted together by any benchmark above.

Evaluation remains a central problem in spoken QA. Exact Match~(EM) and F1 have been default since SQuAD~\cite{squad}, but are brittle to paraphrase~\cite{bertmatching,li_pedants_2024}, especially where ASR errors and bilingual transcriptions produce correct answers that don't match exactly. Model-based metrics such as Bert Matching~\cite{bertmatching}, BERTScore~\cite{zhang2019bertscore}, BLEURT~\cite{sellam2020bleurt}, and ORCA~\cite{sedlacek-etal-2026-orca} improve on lexical overlap, but are primarily trained for English and cannot be directly applied to Telugu without language-specific fine-tuning, which is outside the scope of this work. This leaves LLM-as-a-judge~\cite{zheng2023judging,fromgenerationtojudgment,llmsasjudgessurvey} and multilingual embedding-based metrics such as BLASER-2.0~\cite{blaser} as the practical options for Telugu SQA evaluation. It is also worth noting that the majority of Indic QA datasets employ automatic evaluation metrics such as EM and F1, and none of them have explored the reliability of LLM-as-a-judge for evaluation. Moreover, prior works~\cite{multilingual-judge,padarha-etal-2025-evaluating} have shown LLM judges are inconsistent across languages and tasks.
We address this gap within the V\={a}kQA benchmark as we quantify the reliability of LLM-as-a-judge for Telugu spoken QA.
We make the following contributions:
\begin{itemize}
    \item We construct and publicly release V\={a}kQA\footnote{\texttt{\url{https://hf.co/datasets/Bhavanaakkiraju/VakQA}}}, to the best of our knowledge the first benchmark for SQA in Telugu across six domains, including spoken audio and bilingual transcriptions.

    \item We analyze the reliability of LLM-based automatic evaluation for Telugu QA and show that it depends strongly on judge model choice, with Gemini-as-judge exhibiting non-uniform strictness, highlighting limitations of current evaluation practice for Telugu QA.

    \item We conduct a systematic benchmark study under different input conditions, isolating the effects of input language, input modality, model size (in parameters), and tier (proprietary vs.\ open-weights), and analyzing cascaded ASR$\rightarrow$MT error compounding and domain-wise QA model performance.

\end{itemize}

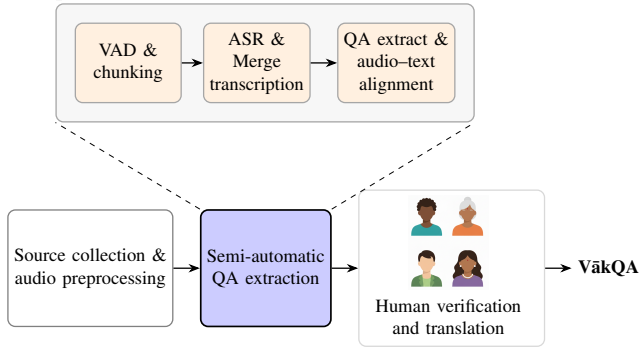
\begin{figure}[!ht]
\centering
\scalebox{0.7}{
\begin{tikzpicture}[node distance=1.3cm]

  \node (input)  [textbox]  {Source collection \&\\ audio preprocessing};

  \node (model)  [llm, right=0.5cm of input]
  {Semi-automatic\\ QA extraction};

  \node (humans) [right=0.5cm of model, draw=gray!30, line width=0.7pt, rounded corners, inner sep=3pt, align=center]
    {
      \begin{tabular}{c}
        \includegraphics[width=1.5cm]{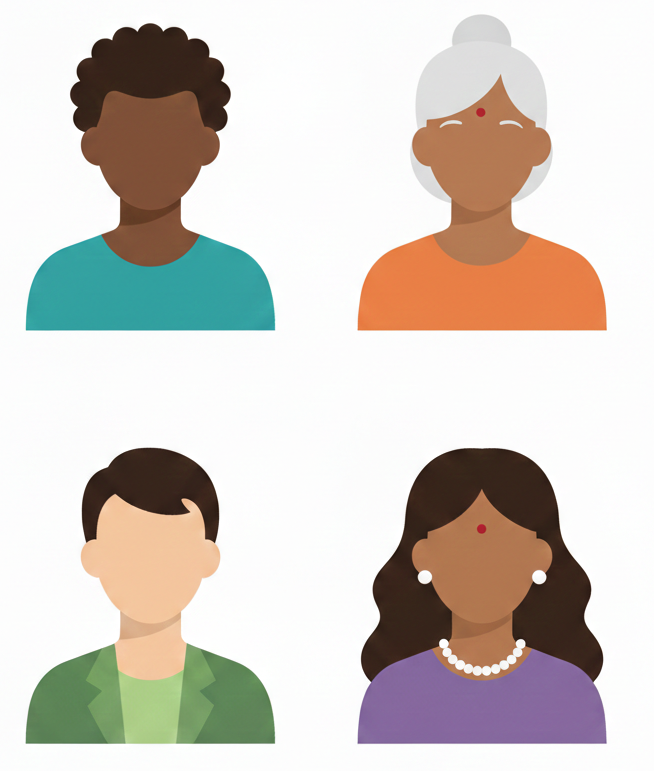}\\
        {Human verification}\\
        {and translation}
      \end{tabular}
    };

  \node (qa)[right=0.5cm of humans, align=left,font=\bfseries] {V\=akQA};

  \draw[arrow] (input) -- (model);
  \draw[arrow] (model) -- (humans);
  \draw[arrow] (humans) -- (qa);

  \node (vad)   [topbox, above=2cm of model, xshift=-2.6cm]
  {VAD \&\\ chunking};

  \node (asr)   [topbox, right=0.4cm of vad] {ASR \&\\ Merge \\ transcription};

  \node (align) [topbox,right=0.7cm of asr, xshift=-0.2cm]
  {QA extract \&\\ audio--text \\ alignment};

  \draw[arrow] (vad) -- (asr);
  \draw[arrow] (asr) -- (align);

  \begin{pgfonlayer}{background}
    \node[groupbox, fit=(vad) (asr) (align)] (topgroup) {};
  \end{pgfonlayer}

  \node[anchor=south west, font=\small, fill=gray!6, inner sep=1.5pt]
    at ([yshift=1pt]topgroup.north west) {};

  \draw[dashed]
     (model.north west) --(topgroup.south west);

  \draw[dashed]
     (model.north east) -- (topgroup.south east);


\end{tikzpicture}
}
\caption{Semi-automatic V\={a}kQA benchmark creation pipeline.}
\label{fig:creation}
\end{figure}

\section{V\={a}kQA Benchmark}
We construct the V\={a}kQA dataset using a multi-step pipeline: (1) data collection and audio extraction, (2) QA pair extraction, and (3) human validation and translation. Figure~\ref{fig:creation} provides an overview of the full workflow.

\subsection{Data Collection and Audio Extraction}
We collected Telugu YouTube videos from channels featuring quiz-style and multiple-choice question answering (MCQ) content, ensuring that each spoken question and its corresponding answer were clearly separated into distinguishable audio segments. To ensure domain diversity, we curated sources across six categories: General Knowledge~(GK), Science, Geography, History, Politics, and Culture.

\subsection{Semi-Automatic QA Pair Extraction}
\label{sec:semi_auto_extraction}
We designed a pipeline to extract candidate QA pairs along with their corresponding audio spans. First, Pyannote VAD~\cite{Bredin23} detects non-silent regions and segments them into 7-second chunks with 2-second overlap, preserving the original timestamps. Each chunk is then transcribed using a fine-tuned Seamless-large-v2 (Seamless FT)~\cite{seamlessm4t2023} Telugu ASR model, trained on approximately 900 hours of data from IndicVoices~\cite{indicvoices}, Kathbath~\cite{kathbath}, Google FLEURS~\cite{fleurs}, SyspIn TTS~\cite{syspin}, and IndicTTS resources~\cite{indictts}. The chunk transcripts are merged into a single passage, and then Gemini is used to extract the QA pairs verbatim. To obtain finer-grained alignment, word-level timestamps are then separately computed using Whisper-timestamped~\cite{radford2022robust,lintoai2023whispertimestamped} with IndicWhisper~\cite{bhogale23_interspeech}. Finally, the extracted QA text is aligned with this word-level ASR output via fuzzy string matching (Levenshtein ratio $\geq$ 85\%), enabling precise adjustment of the QA audio segment boundaries.

\subsection{Human Verification and Translation}


Five annotators verified the extracted QA pairs by checking whether the question and answer transcripts matched their corresponding audio segments. Annotators then manually translated all Telugu QA pairs into English, producing bilingual question--answer pairs. The resulting dataset comprises 2,001 spoken questions with a total audio duration of 2.53 hours spanning six domains: Science (27\%), General Knowledge (23\%), Politics (16\%), History (13\%), Culture (12\%), and Geography (10\%). The statistics are given in Table~\ref{tab:dataset_stats}. A small number of instances (6 in Telugu and 8 in English) contain long descriptive answers, which were retained as-is despite their potential impact on EM and F1 metrics.




\begin{table}[!t]
\centering
\caption{Dataset statistics for the V\=akQA benchmark.}
\label{tab:dataset_stats}
\begin{tabular}{lrr}
\toprule
\textbf{Statistic} & \textbf{Telugu} & \textbf{English} \\
\midrule
Total questions & {2,001} & 2,001\\
Total audio duration (hrs) & {2.53} & --\\
Avg.\ audio length (sec) & {4.55} & --\\
\midrule
Avg.\ words per question & 7.2 & 9.9 \\
Avg.\ words per answer & 2.4 & 2.4 \\
\bottomrule
\end{tabular}
\end{table}


\section{Experimental Setup}
\label{sec:exp_setup}

\subsection{QA Models and Input Configurations}
\label{ssec:qa_models}
We evaluate two categories of QA models: (i) a proprietary model accepting speech or text input, and (ii) open-weight text-only models. To the best of our knowledge, no open-weight Telugu speech LLM is currently available. We use Gemini-2.5-Flash (Gemini) as the proprietary model. For open-weight LLMs, we consider Gemma-3 family (4B, 12B, 27B) of models~\cite{gemmateam2025gemma3technicalreport}, Llama-3.1~\cite{grattafiori2024llama3herdmodels}, Hex-1~\cite{hex}, Sarvam-m~\cite{sarvam}, and Qwen-3-4B~\cite{qwen3technicalreport}.
\subsubsection*{\textbf{Input modality and language}}

We evaluate the QA models across two dimensions: \emph{input modality} (speech or text) and \emph{input language} (Telugu or English). In the \textbf{direct speech} setting, raw Telugu audio is provided to Gemini; open-weight models are not evaluated here as they do not accept Telugu speech input. For \textbf{Telugu ASR text}, speech is transcribed using either Seamless FT or IndicWhisper~\cite{indicvoices} and fed to the QA models. In the \textbf{cascaded ASR$\rightarrow$MT (English)} setting, Telugu ASR transcripts are translated into English using either Seamless MT or Indic MT~\cite{gala2023indictrans2}, yielding four ASR$\rightarrow$MT configurations (2 ASR systems and 2 MT systems). Finally, the \textbf{oracle text} setting uses ground-truth Telugu and English text to isolate the impact of ASR and MT errors. Testing the same QA model with both languages on identical questions allows us to distinguish between two failure modes: a model lacking knowledge entirely versus one that has knowledge but cannot access it in one or the other language.



\subsection{Evaluation Metrics}
We evaluate the answer correctness of QA models using human judgments and automatic metrics, with human ratings serving as the gold-standard reference. For automatic evaluation, we report Exact Match (EM), token-level F1, BLASER-2.0, and an LLM-as-a-judge correctness score.

\subsubsection*{\textbf{Human judgments}}
We sampled 100 questions in Telugu textual form and obtained candidate answers from four QA models: Gemini, and Gemma-3 (4B, 12B, 27B) variants, producing 400 \emph{(question, reference answer, candidate answer)} triplets. Five native Telugu speakers (including co-authors) assigned human ratings to each candidate answer on a 1--5 scale following the rubric given in Table~\ref{tab:rubric}.  Inter-rater reliability was measured using Krippendorff's $\alpha$~\cite{krippendorff_content_2019}. After excluding 20 outlier items with unusually inconsistent ratings (i.e., items where annotator scores spanned the full 1--5 range), $\alpha$ reached 0.836, indicating strong agreement and supporting the reliability of our human annotations. These annotations were used as the gold-standard reference to assess the reliability of automatic evaluation metrics.

\begin{table}[!t]
\centering
\caption{Rubric for evaluating answer correctness}
\label{tab:rubric}
\begin{tabular}{@{} c p{0.8\columnwidth} @{}} 
\toprule
\textbf{Score} & \textbf{Description} \\
\midrule
\textbf{1} & \textbf{Irrelevant:} The answer does not address the question or is completely off-topic. \\
\addlinespace[0.5em]
\textbf{2} & \textbf{Poor:} The answer addresses the question but contains major inaccuracies or misunderstandings. \\
\addlinespace[0.5em]
\textbf{3} & \textbf{Fair:} The answer is partially correct but misses key details or includes minor errors. \\
\addlinespace[0.5em]
\textbf{4} & \textbf{Mostly Correct:} The answer is largely accurate with only small errors or omissions. \\
\addlinespace[0.5em]
\textbf{5} & \textbf{Fully Correct:} The answer is completely accurate and fully addresses the question. \\
\bottomrule
\end{tabular}
\end{table}


\subsubsection*{\textbf{LLM-as-a-judge}}
We use the proprietary model Gemini and three variants of Gemma-3 as judges. Each judge scores a model-generated answer by comparing it against the question and reference answer using the same 1--5 rubric as human evaluation. The evaluation prompt was iteratively refined to maximize correlation with human ratings.


\subsubsection*{\textbf{Lexical and Embedding-based Metrics}}
We use EM, token-level F1 as lexical and BLASER-2.0 as embedding-based metrics. EM and F1 measure exact string match and token overlap respectively, while BLASER-2.0 is a sentence-level embedding-based semantic similarity metric producing scores on a 1--5 scale.

%
%


%

\section{Evaluation Reliability}
\label{sec:eval_reliability}

We first present results on the evaluation reliability, followed by the analysis of QA systems with varying configurations.

\begin{table}[!t]
\centering
\caption{Comparison of LLM-as-Judge models, lexical and embedding-based metrics against human judgments.}
\begin{tabular}{llcccc}
\toprule
\textbf{Type} & \textbf{Name} & \textbf{$\rho$ }$\uparrow$ & \textbf{$\tau$} $\uparrow$ & \textbf{ME} $\downarrow_0$ & \textbf{MAE} $\downarrow$\\
\midrule
\multirow{4}{*}{LLM Judge} & Gemma-3-4B & 0.57 & 0.50 & 0.31 & 0.87\\
& Gemma-3-12B & 0.81 & 0.71 & 0.34 & 0.62\\
& Gemma-3-27B & 0.80 & 0.70 & \textbf{-0.07} & 0.55\\
& Gemini & \textbf{0.86} & \textbf{0.77} & -0.28 &  \textbf{0.46}\\
\midrule
\multirow{2}{*}{Lexical}& EM & 0.37 & 0.33 & -2.47 &2.47 \\
& Avg.\ F1 & 0.49 & 0.43 & -2.40 & 2.40 \\
\midrule
Embedding-based& BLASER-2.0 & 0.36 & 0.27 & 0.36 & 1.35 \\
\bottomrule
\end{tabular}
\label{tab:correlation}
\end{table}

\begin{figure}[t]
    \centering
    \includegraphics[width=0.5\textwidth]{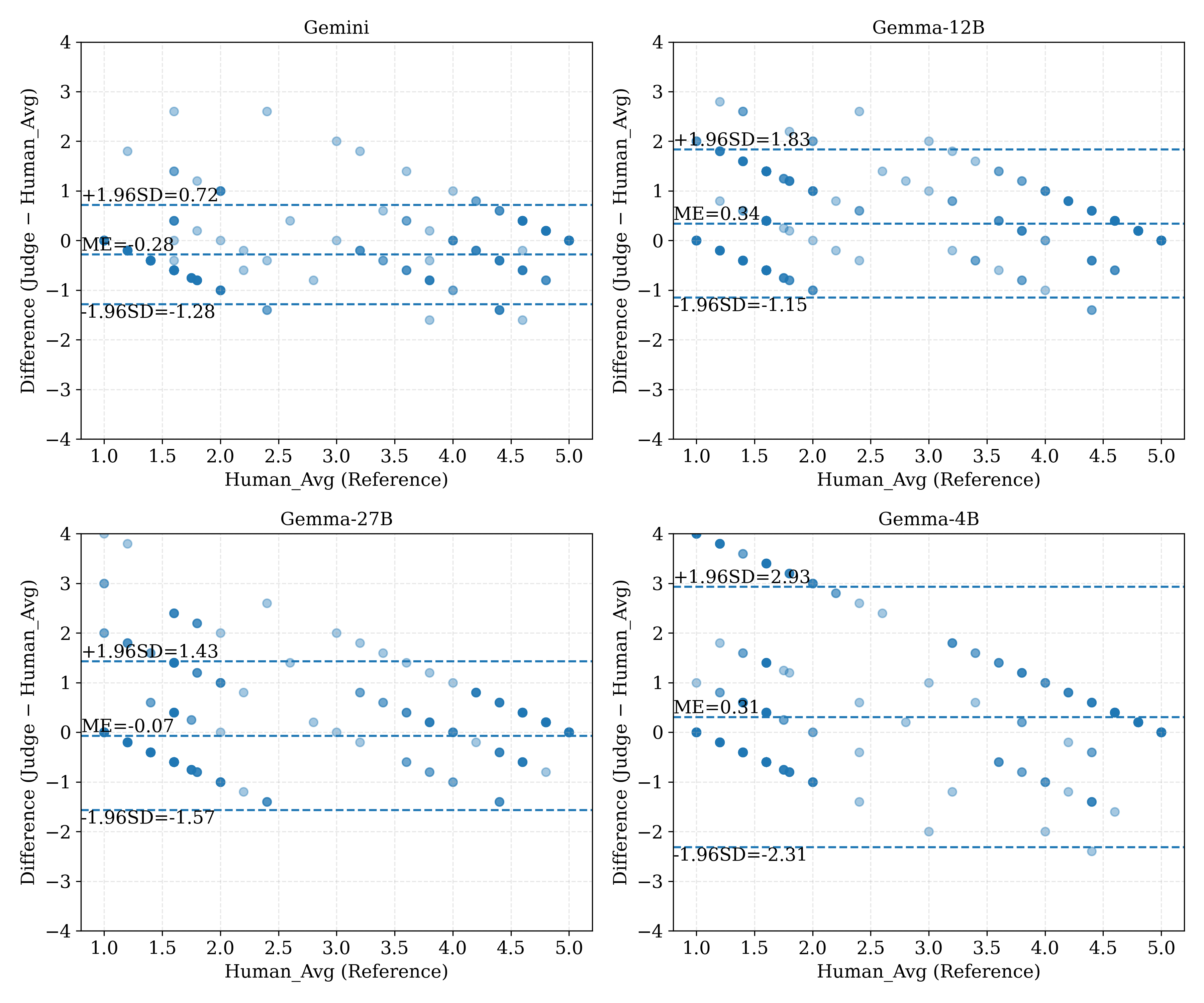}
    \caption{
    Analysis comparing human average scores with LLM-as-judge scores.
    The x-axis represents the average human score (reference), and the y-axis shows the difference between judge and human scores.
    The dashed lines indicate the mean error (bias) and the limits of agreement ($\pm 1.96$ SD).
    }
    \label{fig:bland_altman_judges}
\end{figure}

We measure the reliability of all the considered evaluation methods by comparing them against average human judgment using Spearman's $\rho$, Kendall's $\tau$, mean error (ME), and mean absolute error (MAE). As shown in Table~\ref{tab:correlation}, Gemini-as-a-judge achieves the highest correlation ($\rho$ = 0.86, $\tau$ = 0.77), outperforming other open-weight models. Among Gemma-3 variants, the 12B ($\rho$ = 0.81, $\tau$ = 0.71) and 27B ($\rho$ = 0.80, $\tau$ = 0.70) variants show better alignment, while the 4B variant is the weakest ($\rho$ = 0.57, $\tau=0.5$). The lexical metrics EM, F1 and the embedding-based metric BLASER-2.0 show much lower correlations ($<0.5)$ with human judgments. An example illustrating the limitations of EM and F1 is given in Table~\ref{tab:examples} row E1: the reference answer indicates the broad region affected by cyclones, whereas the candidate answer lists the specific states within that region. Although this is correct and more detailed, EM and F1 assign 0 due to low lexical overlap, while BLASER-2.0 yields a moderate score of 2.43. While Gemini is the most reliable judge in our findings, it is not perfectly aligned with human ratings. Figure~\ref{fig:bland_altman_judges} shows that Gemini is slightly stricter on average (ME=-0.28) with the narrowest limits of agreement (LoA: -1.28 to 0.72), though its behavior is non-uniform: more lenient for low-quality answers and stricter for high-quality ones. Gemma judges exhibit wider LoA: Gemma-12B shows positive bias (ME=0.34; LoA: -1.15 to 1.83); Gemma-27B shows near-zero bias (ME =-0.07; LoA: -1.57 to 1.43); and Gemma-4B shows the largest spread (LoA: -2.31 to 2.93).

\begin{table*}[t]
  \caption{Examples illustrating effect of various aspects: evaluation metrics, and input configurations. All candidate answers are generated by Gemini QA model. Native Telugu text is transliterated into Latin using ISO-15919 standard to improve readability for wider audience. \textit{transl:} is the English translation for the Telugu text provided for the reader---these are not fed as input to the model. Refer to the text from Sections~\ref{sec:eval_reliability} and~\ref{sec:benchmarking_results} for detailed explanations.}
\label{tab:examples}
\centering
\renewcommand{\arraystretch}{1.6}
\setlength{\tabcolsep}{6pt}
\begin{tabular}{llp{5.8cm}p{3.4cm}p{3.8cm}}
\toprule
\textbf{ID} & \textbf{Aspect} & \textbf{Question} & \textbf{Reference Answer} & \textbf{Candidate Answer} \\
\midrule
E1 & Evaluation &
  bh\={a}ratad\={e}\'{s}a\.{m}l\={o} taracug\={a} tuph\={a}nulaku gurayy\={e} pr\={a}\.{m}t\={a}lu \={e}vi? (\textit{transl: Which regions in India are frequently affected by cyclones?}) &
  t\={u}rpu t\={i}ra pr\={a}\.{m}t\={a}lu (\textit{transl: The East coastal regions}) &
  \={a}\.{m}dhraprad\={e}\'{s}, o\d{d}i\'{s}\={a}, pa\'{s}cima be\.{m}g\={a}l, tami\d{l}an\={a}\d{d}u   
  (\textit{transl: Andhra Pradesh, Odisha, West Bengal, Tamil Nadu}) \\
\midrule
E2 & LLM judges &
prapa\.{m}ca v\={a}t\={a}vara\d{n}a din\={o}tsav\={a}nni \={e} r\={o}juna jarupuku\.{m}\d{t}\={a}ru? (\textit{transl: On which day is World Meteorological Day celebrated?}) &
m\={a}rci iravai m\={u}\d{d}u
(\textit{transl: March twenty-three}) &
m\={a}rci 23 (\textit{transl: March 23}) \\
\midrule
\multirow{2}{*}{E3} & Oracle Telugu text &
 mana d\={e}\'{s}apu mo\d{t}\d{t}amoda\d{t}i upagraha\.{m} \={e}di? &
 \={a}ryabha\d{t}\d{t}a  (\textit{Aryabhata}) &  \={a}ryabha\d{t}\d{t}a
 (\textit{Aryabhata}) \\
 & Oracle English text &
  What was our country's first satellite? &   Aryabhata &   Sputnik \\
\midrule
\multirow{2}{*}{E4} & Text Input &
tela\.{m}g\={a}\d{n}a r\={a}\d{s}\d{t}r\={i}ya pa\.{m}\d{d}u \={e}mi\d{t}i?
\textit{(transl: What is the state fruit of Telangana?)} & m\={a}mi\d{d}i (\textit{transl: Mango}) &
m\={a}mi\d{d}i (\textit{transl: Mango}) \\
 & Speech Input &
 &
m\={a}mi\d{d}i (\textit{transl: Mango}) &
batukamma (\textit{Bathukamma}) \\
\midrule
\multirow{2}{*}{E5} & Oracle Telugu text &
t\={a}mara v\={e}\d{t}i valla kalugutu\.{m}di? (\textit{transl: What causes ringworm?}) &
pha\.{m}gas (\textit{Fungus}) & pha\.{m}gas (\textit{Fungus}) \\
& ASR text & s\={a}mar v\={e}\d{t}a valla kalugutu\.{m}di? (\textit{transl: Is Samar caused by hunting?}) &
pha\.{m}gas (\textit{Fungus}) & v\={e}\d{d}imi (\textit{transl: heat}) \\
\midrule
E6 & Domain: Culture &
  \d{r}gv\={e}da\.{m}l\={o} kulapa a\.{m}\d{t}\={e} evaru?
 (\textit{transl: Who is referred to as ``Kulapa'' in the Rigveda?}) &
 ku\d{t}u\.{m}ba pedda
 (\textit{transl: head of the family}) &
 ku\d{t}u\.{m}ba pedda
 (\textit{transl: head of the family}) \\
\midrule
E7 & Domain: Science &
  n\={e}la guri\.{m}ci adhy\={a}yana\.{m} c\={e}s\={e} \'{s}\={a}stra\.{m} \={e}di?
 (\textit{transl: What is the study of soil called?}) &
  pe\d{d}\={a}laj\={i}
 (\textit{transl: Pedology}) &
  bh\={u}garbha\'{s}\={a}stra\.{m}
 (\textit{transl: Geology}) \\
\bottomrule
\end{tabular}
\end{table*}

\subsubsection*{\textbf{Pairwise comparison of judge models}}
We further examine the sensitivity of scores to the judge model by doing pairwise comparison of LLM-judgements for each of the 2,001 candidate answers from Gemini QA model. By swtiching the judge from Gemini to Gemma-12B we observed that 46.23\% of candidate answers received worse scores, while only 21.14\% improved and 32.63\% remain unchanged (row 1 in Table~\ref{tab:pairwise}). Row E2 from Table~\ref{tab:examples} presents an example where the candidate answer is correct; Gemini-as-a-judge rates it correctly at 5, however Gemma-3-12B rates it as 1, due to sensitivity to surface-form variation. This failure suggests that Gemma-3-12B struggles to recognize semantic equivalence in Telugu (e.g., numeric vs.\ spelled-out dates, or correct short forms when the reference includes glosses). Overall, while the proprietary QA model performs substantially better than open-weight models, reliable evaluation in this low-resource setting also requires a sufficiently capable judge. Based on these results, Gemini is used as the primary evaluation method for the results presented in the subsequent sections.

\section{Benchmarking QA Models}
\label{sec:benchmarking_results}

\begin{table*}[!ht]
\centering
\caption{Evaluation of QA models across input configurations using Gemini-as-a-judge on 1--5 scale (mean (std)). G27B/G12B/G4B represents Gemma-3 (27B/12B/4B); L3-8B: Llama-3.1-8B; Q3-4B: Qwen-3-4B; Srv-m: Sarvam-m.}
{
\begin{tabular}{lllcccccccc}
\toprule
\textbf{ID} & \textbf{Input} & \textbf{Lang} &
\multicolumn{8}{c}{\textbf{QA Models}} \\
\cmidrule(lr){4-11}
& & & \textbf{Gemini} & \textbf{G27B} & \textbf{G12B} & \textbf{G4B} & \textbf{L3-8B} & \textbf{Hex-1} & \textbf{Q3-4B} & \textbf{Srv-m} \\
\midrule

S1 & Speech & Te
& 3.28 (1.84) & -- & -- & -- & -- & -- & -- &-- \\

\midrule

O1 & Oracle Text & Te
& \textbf{3.63 (1.71)} & 2.55 (1.79) & 2.01 (1.60) & 1.43 (1.12) & 1.41 (1.11) & 1.33 (1.0) & 1.17 (0.72) & 1.98 (1.60)\\

O2 & Oracle Text & En
& 3.52 (1.74) & -- & -- & -- & -- & -- & -- & --\\

\midrule

A1 & Seamless FT ASR & Te
& 3.40 (1.79) & 2.34 (1.76) & 1.84 (1.51) & 1.38 (1.06) & -- & -- & -- & 1.75 (1.45)\\

A2 & IndicWhisper ASR & Te
& 3.09 (1.85) & 2.22 (1.72) & 1.77 (1.47) & 1.33 (0.99) & -- & -- & --& 1.69 (1.39)\\

\midrule

O3 & O1 $\rightarrow$ Seamless MT & En
& 2.74 (1.83) & 2.45 (1.79) & 2.28 (1.73) & 1.84 (1.51) & 2.08 (1.67) &1.96 (1.59) & 1.74 (1.46) & 2.29 (1.76)\\

O4 & O1 $\rightarrow$ Indic MT & En
& 3.05 (1.85) & 2.67 (1.82) & 2.38 (1.75) & 1.96 (1.59) & 2.21 (1.73) & 2.06 (1.64) & 1.84 (1.53) & 2.39 (1.76)\\

\midrule

M1 & A1 $\rightarrow$ Seamless MT & En
& 2.51 (1.82) & 2.28 (1.77) & 2.06 (1.64) & 1.70 (1.43) & 1.87 (1.57) & -- & --& 2.1(1.66 ) \\

M2 & A1 $\rightarrow$ Indic MT & En
& 2.80 (1.87) & 2.45 (1.81) & 2.13 (1.67) & 1.79 (1.49) & 1.96 (1.61) & -- & -- & 2.18 (1.7)\\

M3 & A2 $\rightarrow$ Seamless MT & En
& 2.44 (1.82) & 2.12 (1.70) & 1.94 (1.58) & 1.62 (1.36) & 1.76 (1.49) & -- & -- & 1.98 (1.6)\\

M4 & A2 $\rightarrow$ Indic MT & En
& 2.61 (1.85) & 2.29 (1.76) & 2.03 (1.63) & 1.67 (1.41) & 1.83 (1.53) & -- & -- & 2.05 (1.65)\\
\bottomrule
\end{tabular}
}
\label{tab:results}
\end{table*}

\begin{table}[!t]
\centering
\caption{Pairwise comparison of pipeline variations against the oracle Telugu baseline (O1). \textbf{Same}: the score difference is zero; \textbf{Better}: the variant scores higher than O1; \textbf{Worse}: the variant scores lower than O1. Each row shows the percentage of questions falling into each category.}
\scalebox{0.9}{
\begin{tabular}{llllrrr}
\toprule
& \multicolumn{3}{c}{\textbf{Configuration}} & \multicolumn{3}{c}{\textbf{vs.\ O1 Gemini QA (\%)}} \\
\cmidrule(lr){2-4} \cmidrule(lr){5-7}
\textbf{Factor} &\textbf{ID} & \textbf{QA model} & \textbf{Judge}
  & \textbf{Same} & \textbf{Better} & \textbf{Worse} \\
\midrule
{Judge} & O1 & Gemini & Gemma-12B & 32.6 & 21.1 & 46.2 \\
\midrule
{Input (ASR)}& A1 & Gemini & Gemini & 83.3 & 4.8 & 11.9 \\
\midrule
{Language} & O2 & Gemini & Gemini & 64.7 & 16.5 & 18.8 \\
\midrule
 Modality & S1 & Gemini & Gemini & 65.7 & 13.1 & 21.2 \\ 
\bottomrule
\end{tabular}
}
\label{tab:pairwise}
\end{table}

We present the QA results across several models highlighting the effects of input language (Telugu vs English), input modality (speech, text), and cascaded pipeline errors.

\subsection{Proprietary vs Open-Weight Models}
Table~\ref{tab:results} shows our main results on V\={a}kQA benchmark. We can observe a consistent gap between Gemini and open-weight models across all input configurations. Best scores are achieved with oracle Telugu text as input (O1): Gemini achieves 3.63~(1.71), while open-weight models score lower: Gemma-27B (2.55~(1.79)), Gemma-12B (2.01~(1.60)), Sarvam-m (1.98~(1.60)), and Gemma-4B (1.43~(1.11)); other models (e.g., Llama-3.1, Hex-1, Qwen-3-4B) score near or below 1.5. Changing the input from oracle text to Telugu ASR transcripts (rows A1/A2) decreases scores across all models but, the gap remains (Gemini: 3.40/3.09; Gemma-27B: 2.34/2.22; Gemma-12B: 1.84/1.77). With translated English inputs (rows O3/O4 and M1--M4), scores drop further due to MT error propagation, though larger open-weight models remain stronger than smaller ones.


\subsection{Effect of Input Language}
As shown in Table~\ref{tab:results}, row O1 (oracle Telugu text) with Gemini QA model achieves a score of 3.63~(1.71), while O2 (oracle English text) scores 3.52~(1.74). To isolate the effect of input language, we do pairwise comparison of O1 and O2 and the results are presented in Table~\ref{tab:pairwise}---relative to O1, switching from Telugu to English input degrades performance of Gemini QA model on 18.8\% of questions, improves it on 16.5\%, and leaves 64.7\% unchanged. This degradation can be attributed in part to ambiguities introduced during translation. For example, row E3 from Table~\ref{tab:examples} shows the same input question in Telugu and English, respectively. In Telugu, a possessive pronoun ``mana (\textit{transl: our})'' implicitly refers to India, making the question's scope clear, and the model correctly answers ``Aryabhata". Once translated into English, this reference becomes ambiguous, and the model treats it as globally scoped, answering ``Sputnik'' instead.

\subsection{Effect of Input Modality}
We next compare the two input modalities: text (O1) vs speech (S1) while keeping the QA model restricted to Gemini. From Table~\ref{tab:results}, we can see that O1 with Gemini QA model achieves a score of 3.63~(1.71), while with S1 it scores 3.28~(1.84). Pairwise comparisons from Table~\ref{tab:pairwise} show that---relative to O1, speech input degrades performance for 21.2\% of questions, improves it for 13.1\%, and leaves 65.7\% unchanged. This drop can be partly explained by acoustic confusions in the speech input. Row E4 from Table~\ref{tab:examples} illustrates the phenomenon---the same question asked both in textual and spoken form to Gemini QA model. The question is about the state ``fruit'' of Telangana. With oracle text in Telugu as input, the model correctly answers with \textbf{mango}. With speech input, the model instead answers \textbf{Bathukamma}, which is one of state ``festivals'' of Telangana. Here, the model appears to confuse the Telugu word ``pa\.{m}du (\textit{transl: fruit})'' for the phonetically closer word ``pa\.{m}\d{d}uga \textit{(transl: festival)}" resulting in the correct festival name instead of the fruit name. This example illustrates how acoustic confusions in speech input can lead to semantic drift in downstream QA.

\begin{table}[!t]
\centering
\caption{ASR and MT results on V\={a}kQA.}
\label{tab:asr_mt}
\begin{tabular}{lrr}
\toprule
\textbf{System} & \textbf{WER$\downarrow$} & \textbf{CER$\downarrow$} \\
\midrule
Seamless FT ASR  & 30.25 & 10.12 \\
IndicWhisper ASR & 35.23 & 22.38 \\
\midrule
             & \textbf{BLEU$\uparrow$} & \textbf{ChrF++$\uparrow$} \\
\midrule
Oracle text $\rightarrow$ Seamless MT  & 37.22 & 60.12 \\
Oracle text $\rightarrow$ Indic MT     & 36.91 & 61.55 \\
Seamless FT ASR $\rightarrow$ Seamless MT & 32.18 & 55.54 \\
Seamless FT ASR $\rightarrow$ Indic MT & 29.10 & 56.18 \\
IndicWhisper $\rightarrow$ Seamless MT & 27.95 & 52.03  \\
\bottomrule
\end{tabular}
\end{table}

\subsection{Effect of Cascaded ASR $\rightarrow$ MT Errors}
To study error propagation in cascaded ASR $\rightarrow$ MT pipelines, we compare systems that introduce ASR and/or MT components against oracle text input baselines. Table~\ref{tab:asr_mt} summarises the component-level ASR and MT scores on V\={a}kQA. When ASR and MT are cascaded, translation quality drops substantially. These pipeline errors carry into QA: The \textit{Input (ASR)} row from Table~\ref{tab:pairwise} shows that using transcripts from Seamless FT ASR causes 11.9\% of questions to perform worse than O1 (oracle text transcripts), with only 4.8\% improving. Row E5 from Table~\ref{tab:examples} illustrate how ASR errors can change the meaning of the question and derail downstream QA. Seamless FT ASR misrecognizes the question word for \textbf{ringworm} as a phonetically similar but unrelated word, producing a corrupted transcription. With oracle text (O1), the model correctly answers \textbf{fungus}; with the corrupted ASR transcript~(A1), the question becomes ill-posed and the model instead answers \textbf{heat}. This shows how moderate ASR error~(WER 30.25) can lead to complete semantic failure downstream. Table~\ref{tab:results} further shows that cascaded pipelines score lower than oracle baselines: O1$\rightarrow$ Seamless MT (O3) scores 2.74 vs.\ 3.63 for O1 (a $\sim$0.9 drop), and full ASR+MT cascades (M1--M4) degrade further, scoring 2.44--2.80, confirming that errors compound across stages and progressively reduce QA performance.

\subsection{Domain-wise Performance}
\begin{figure}[t]
    \centering
    \includegraphics[width=0.8\linewidth]
    {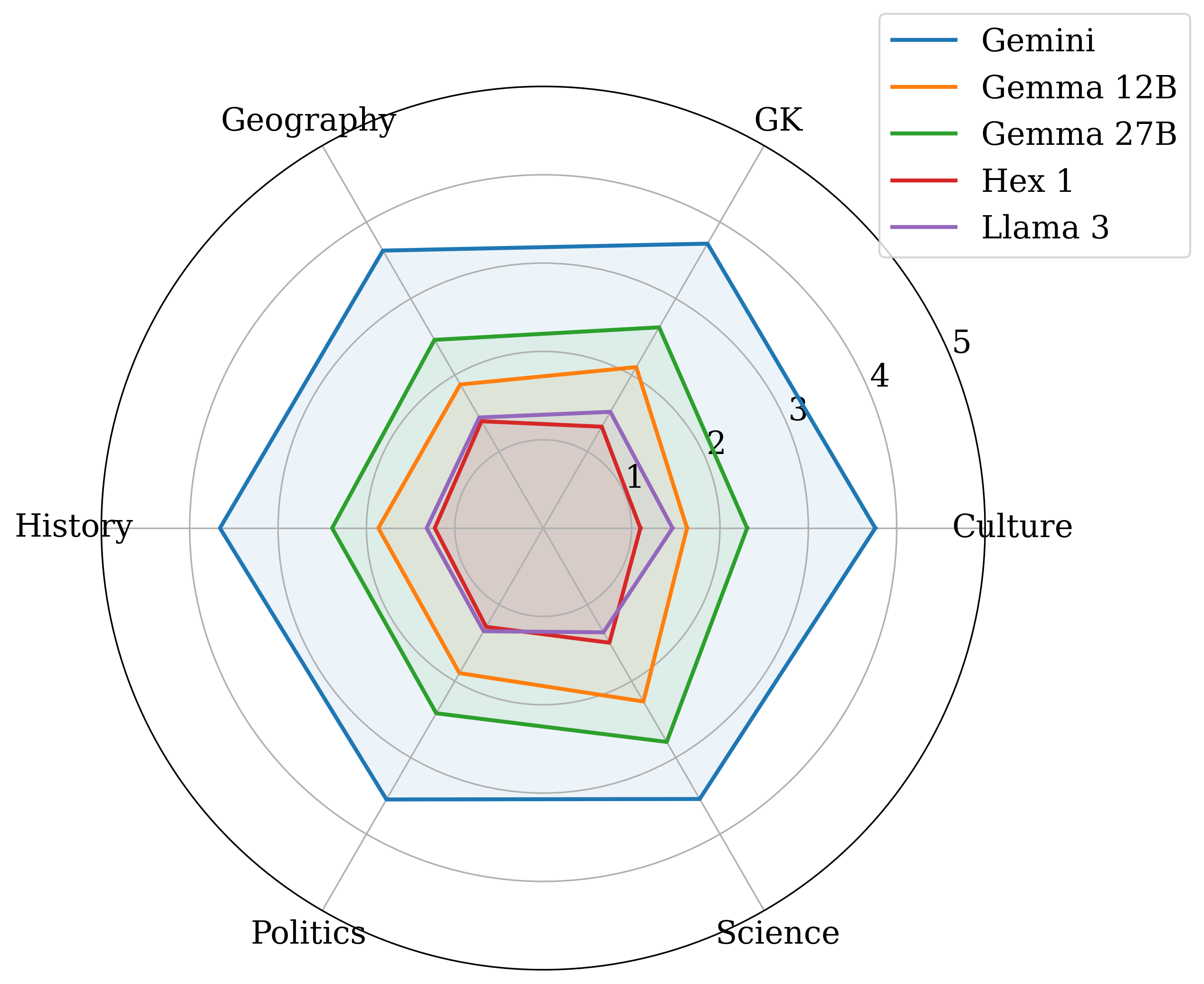}
    \caption{Domain-wise average answer correctness scores for various QA models with oracle Telugu text as input (O1).}
    \label{fig:o1_domain}
\end{figure}

We analyze domain-wise performance under two input configurations: O1 (oracle Telugu text) and O3 (oracle Telugu text $\rightarrow$ Seamless MT). Figures~\ref{fig:o1_domain} and~\ref{fig:o3_domain} show spider plots comparing average answer correctness scores for various QA models across the six domains.

\subsubsection*{\textbf{Gemini performance across domains}}
From Fig.~\ref{fig:o1_domain}, we can see that---with oracle Telugu text as input, Gemini scores consistently across all domains (3.54--3.76); Culture is strongest (3.76), as Telugu input preserves cultural cues (E6, Table~\ref{tab:examples}). Science and Politics are slightly weaker (3.54), often due to specialized terminology (E7, Table~\ref{tab:examples}) where Gemini answers ``geology" instead of ``pedology."

\subsubsection*{\textbf{Cross-model comparison}}
\begin{figure}[t]
    \centering
    \includegraphics[width=0.8\linewidth]{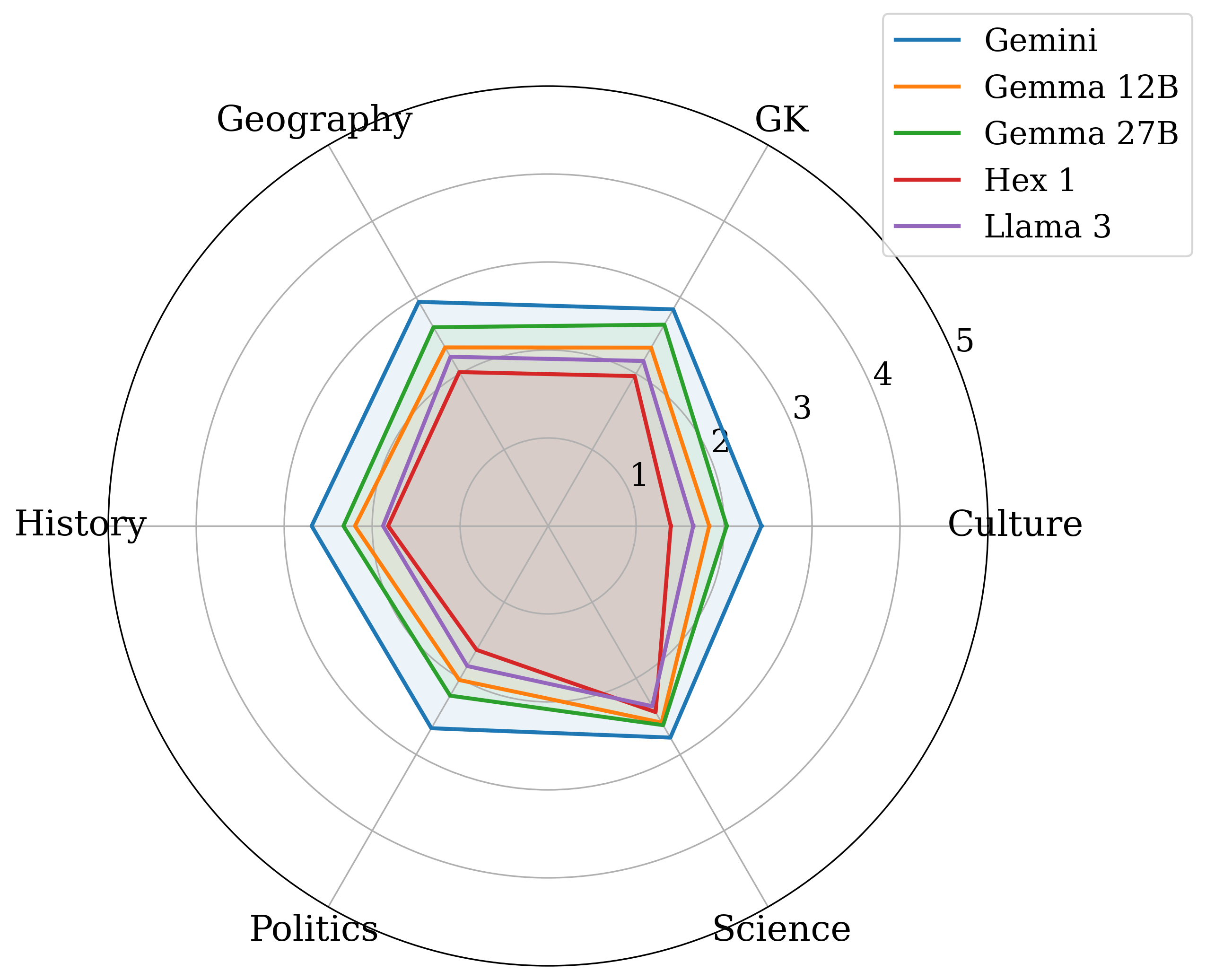}
    \caption{Domain-wise average answer correctness scores for various QA models with machine translated English text as input (O3).}
    \label{fig:o3_domain}
\end{figure}

Comparing Figures~\ref{fig:o1_domain} and \ref{fig:o3_domain} domain-wise we can identify the following patterns. Performance of Gemini QA model in Culture shows the largest decrease (3.76 $\rightarrow$ 2.42), consistent with translation obscuring key details --- for example, the Rigveda question in E6 (Table~\ref{tab:examples}) yielded an incorrect answer once translated. In contrast, Science benefited from English phrasing in some cases: the soil-science question in E7 (Table~\ref{tab:examples}) is correctly answered as ``pedology" once translated into English. Geography becomes the strongest domain in O3~(2.94), suggesting that questions dominated by place names and proper nouns transfer more reliably across languages.
Across models, larger open-weight QA models perform better but still lag behind Gemini. Gemma-27B shows moderate domain variation in O1 (2.31--2.80), while Gemma-12B is lower overall (1.63--2.27). Smaller models such as Hex-1 and Llama-3 perform poorly across domains, often below 1.5. For open-weight models, Science is generally the easiest domain, whereas Culture is consistently the hardest, reflecting the difficulty of culturally specific questions.

\section{Conclusions}
We introduced and released V\={a}kQA, the first benchmark for Telugu spoken factoid
question answering, and showed that it remains challenging for both
proprietary and open-weight models, with proprietary systems performing
consistently better. Performance is shaped by input formulation and pipeline
design: Telugu text better preserves scope and specificity than English
translations, speech input introduces acoustic confusions that alter question
meaning, and cascaded ASR$\rightarrow$MT pipelines compound errors
progressively. Domain also matters --- Culture is the hardest domain for
open-weight models and the most sensitive to translation, while Science and
Geography transfer more reliably across languages due to stable terminology
and proper nouns. Reliable evaluation remains a bottleneck: smaller
open-weight LLM-judges fail to recognize semantic equivalence in Telugu, and
even Gemini-as-a-judge is non-uniform, being more lenient at low scores and
stricter at high scores. Three limitations follow from these findings:
First, YouTube-sourced audio requires faithful transcript-based translation 
rather than clarified rewrites, which introduces scope ambiguity --- as in E3
(Table~\ref{tab:examples}), where a Telugu possessive pronoun becomes
ambiguous in English. Second, Gemini-as-a-judge's non-uniform strictness
limits fine-grained comparisons. Third, some Science and Geography reference
answers use English transliterations, and we do not assess whether judges
score these consistently against native Telugu equivalents.

\section*{Acknowledgments}
Santosh Kesiraju was supported by Ministry of Education, Youth and Sports of the Czech Republic (MoE) through the OP JAK project ``Linguistics, Artificial Intelligence and Language and Speech Technologies: from Research to Applications'' (ID:CZ.02.01.01/00/23\_020/0008518).

\section*{Generative AI Use Disclosure}
Generative AI tools were used to assist with the writing and language editing of this paper.

\bibliographystyle{IEEEtran}
\bibliography{mybib}

\end{document}